\documentclass[runningheads,a4paper]{llncs}
\usepackage{dsfont}
\usepackage{amsmath}
\usepackage{amssymb}
\usepackage{graphicx}
\usepackage{subfigure}
\usepackage{wrapfig}
\usepackage{cite}
\usepackage{bm}
\usepackage{hyperref}

\begin{document}

\mainmatter  

\title{Learning-based Ensemble Average Propagator Estimation}

\titlerunning{LEAPE}

%
%

\author{Chuyang Ye}
%
\authorrunning{C. Ye}

\institute{National Laboratory of Pattern Recognition \& Brainnetome Center, Institute of Automation, Chinese Academy of Sciences, Beijing, China}

%
%

\toctitle{}
\tocauthor{}
\maketitle

\begin{abstract}
By capturing the anisotropic water diffusion in tissue, \textit{diffusion magnetic resonance imaging}~(dMRI) provides a unique tool for noninvasively probing the tissue microstructure and orientation in the human brain. The diffusion profile can be described by the \textit{ensemble average propagator}~(EAP), which is inferred from observed diffusion signals. 
However, accurate EAP estimation using the number of diffusion gradients that is clinically practical can be challenging. 
In this work, we propose a deep learning algorithm for EAP estimation, which is named \textit{learning-based ensemble average propagator estimation}~(LEAPE). The EAP is commonly represented by a basis and its associated coefficients, and here we choose the SHORE basis and design a deep network to estimate the coefficients. The network comprises two cascaded components. The first component is a \textit{multiple layer perceptron} (MLP) that simultaneously predicts the unknown coefficients. However, typical training loss functions, such as mean squared errors, may not properly represent the geometry of the possibly non-Euclidean space of the coefficients, which in particular causes problems for the extraction of directional information from the EAP. Therefore, to regularize the training,
in the second component we compute an auxiliary output of approximated \textit{fiber orientation}~(FO) errors with the aid of a second MLP that is trained separately. 
We performed experiments using dMRI data that resemble clinically achievable $q$-space sampling, and observed promising results compared with the conventional EAP estimation method.

\end{abstract}

\section{Introduction}
\label{sec:intro}

\textit{Diffusion magnetic resonance imaging} (dMRI) enables noninvasive investigation of brain connectivity and tissue microstructure by capturing the anisotropic water diffusion in tissue. The diffusion profile can be described by the \textit{ensemble average propagator}~(EAP), where both directional and scalar properties of tissue can be derived~\cite{johansen}.

Various methods have been proposed to estimate the EAP. For example, \textit{diffusion spectrum imaging}~(DSI) densely samples the $q$-space and takes the inverse Fourier transform of the normalized diffusion signals to compute the EAP. However, DSI uses a large number of measurements, which requires a long acquisition time and is impractical for clinical use. Therefore, to reduce the number of required measurements, methods have been developed to model the diffusion signals with an adequate basis, 
and EAP estimation is equivalent to the estimation of basis coefficients~\cite{Descoteaux,Assemlal,Merlet}. 

Deep learning has been successfully applied to many computer vision tasks~\cite{LeCun}, including the computation of some diffusion properties~\cite{Golkov}. In~\cite{Golkov}, a \textit{multiple layer perceptron}~(MLP) has been used to predict scalar diffusion features, such as tissue microstructure and diffusion kurtosis, and the quality of estimation is improved.
However, extending the MLP structure to estimate the EAP---i.e., a vector of basis coefficients---is not trivial,  because the coefficients could lie in a non-Euclidean space. Typically used training error measures, such as mean squared errors, may not properly define the difference between the estimated and training EAPs, which in practice
causes problems in particular for the extraction of directional information from the EAP.

In this work, we propose a deep learning algorithm for EAP estimation, which is named \textit{learning-based ensemble average propagator estimation}~(LEAPE). 
We select the SHORE basis for demonstration, which allows closed-form computation of some diffusion features~\cite{Merlet}.
LEAPE comprises two cascaded components.
The first component is an MLP that simultaneously predicts the basis coefficients. 
Then, using the output of the first component and the training EAPs, we compute an auxiliary output of \textit{fiber orientation}~(FO) errors in the second component to regularize the training. Since the computation of the FO error is complex and its gradient is hard to compute for training, the FO error is approximated with the aid of a second MLP, which is trained separately. 
The network was trained using dMRI data densely sampling the $q$-space and evaluated on dMRI data resembling clinically achievable $q$-space sampling. Promising results were observed compared with conventional EAP estimation.

\section{Methods}
\label{sec:method}

\subsection{EAP Representation Using the SHORE Basis}

Suppose the diffusion signal associated with a diffusion gradient $\bm{q}_{k}$~$(k=1,\ldots,K)$ is denoted by $S(\bm{q_{k}})$ and the signal without diffusion weighting is $S_{0}$. We call $y_{\bm{q}_{k}}=S(\bm{q_{k}})/S_{0}$ the normalized diffusion signal, which can be continuously modeled using the SHORE basis~\cite{Merlet}
\begin{eqnarray}
y_{\bm{q}_{k}} = \sum_{n=0}^{N}\sum_{l=0}^{n}\sum_{m=-l}^{l}c_{nlm}\Phi_{nlm}(\bm{q}_{k}).
\label{eqn:shore}
\end{eqnarray}
Here, $\Phi_{nlm}$ is the basis element (see~\cite{Merlet} for its expression) with the radial order~$n$, angular order $l$, and angular degree $m$, $c_{nlm}$ is $\Phi_{nlm}$'s coefficient, and $N$ is the maximal radial order. 
The EAP $P(\bm{R})$ with displacement $\bm{R}$ is calculated from the diffusion signal $y_{\bm{q}_{k}}$ using a Fourier transform, which leads to
\begin{eqnarray}
P(\bm{R})= \sum_{n=0}^{N}\sum_{l=0}^{n}\sum_{m=-l}^{l}c_{nlm}\Psi_{nlm}(\bm{R}).
\label{eqn:eap}
\end{eqnarray}
Here, $\Psi_{nlm}(\bm{R})$ is determined by $n$, $l$, $m$, and $\bm{R}$ (see~\cite{Merlet} for its specification).
Therefore, the EAP is known once the coefficients are obtained, and we will use the estimation of EAPs and these coefficients interchangeably.

We can write Eq.~(\ref{eqn:shore}) in a matrix form $\bm{y} = \mathbf{\Phi} \bm{c}$,
where $\bm{y}=(y_{\bm{q}_{1}},\ldots,y_{\bm{q}_{K}})$, $\mathbf{\Phi}$~is a matrix consisting of the values of basis elements in each diffusion gradient, and $\bm{c}$ is the vector of the basis coefficients. Then, $\bm{c}$ can be estimated by solving an $\ell_{1}$- or $\ell_2$-norm regularized least squares problem~\cite{Merlet}.

From the EAP, scalar and directional information about diffusion can be extracted. In particular, the \textit{orientation distribution function}~(ODF), which plays an important role in fiber tract reconstruction, can be obtained using a linear transformation of $\bm{c}$. Briefly, let $\bm{v}$ be the ODF values sampled in a set of discrete directions. We have $\bm{v}=\mathbf{\Upsilon}\bm{c}$,
where $\mathbf{\Upsilon}$ is a constant matrix encoding the sampled directions (see~\cite{Merlet} for its specification). Then, FOs can be extracted by selecting the peak directions in ODFs, and they are key features for fiber tracking and brain connectivity computation.

\subsection{A Deep Network for EAP Estimation}

Deep learning has recently been applied to dMRI analysis, where tissue microstructure is estimated using an MLP with three hidden layers, and improved estimation results have been demonstrated~\cite{Golkov}. 
However, in~\cite{Golkov} estimation of directional information has not been explored, and for each new scalar diffusion property of interest, another deep network needs to be constructed. If the EAP could be directly estimated using a deep network, then all information encoded in the EAP would be readily extracted. 

\begin{figure}[!t]
  \centering
	\includegraphics[width=0.9\columnwidth]{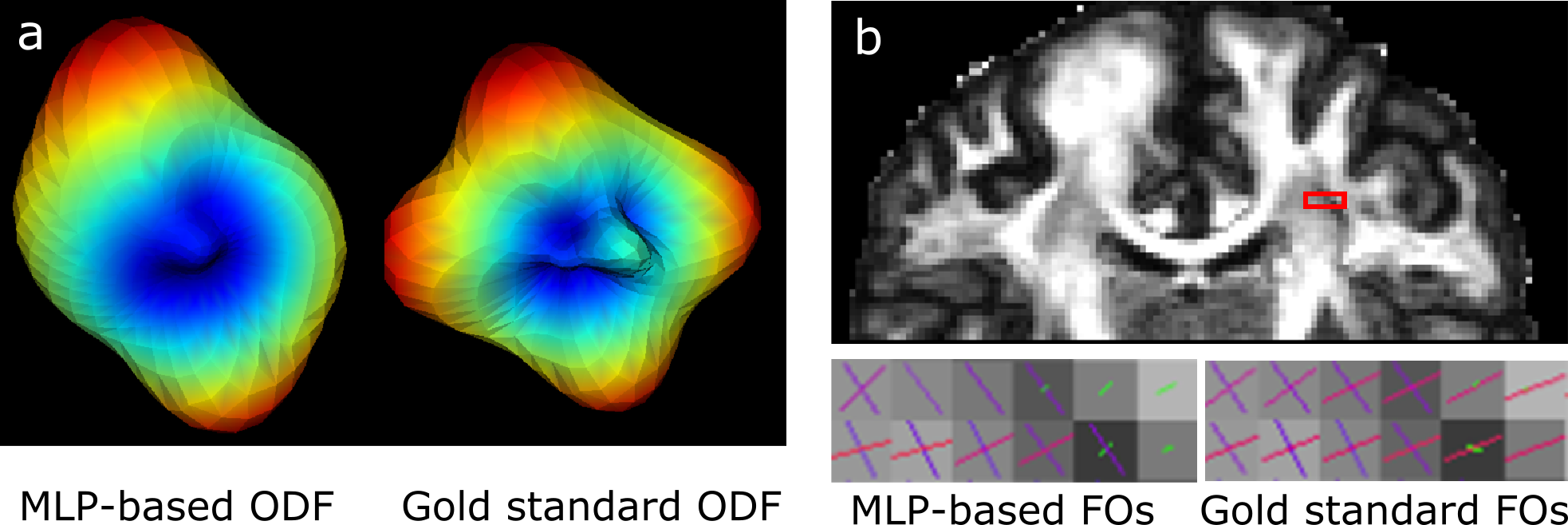}
\caption{An example of the directional information extracted from the EAPs estimated by a simple MLP and gold standard EAPs: (a) ODFs and (b) FOs (overlaid on FA).}
\label{fig:fo_eg}
\end{figure}

A straightforward strategy for EAP estimation would be to modify the MLP in~\cite{Golkov} by giving vectors of coefficients at the output layer. However, typical loss functions, such as the mean squared error used in~\cite{Golkov}, cannot guarantee to properly capture the structure of the coefficients, which could lie in a non-Euclidean space. 
For example, we observed that the extraction of directional information of ODFs and FOs from the EAPs can be problematic with such a strategy. Fig.~\ref{fig:fo_eg} gives an example of the ODF and FOs (overlaid on \textit{fractional anisotropy}~(FA)) extracted from the EAPs estimated using this simple strategy. The horizontal peak of the ODF is not preserved, and incorrect FO configurations with missing FOs can be observed.
Even if a loss function that properly defines the difference of EAP coefficients can be discovered, computing its gradient in the training process can be nontrivial.

\begin{figure}[!t]
  \centering
	\includegraphics[width=0.98\columnwidth]{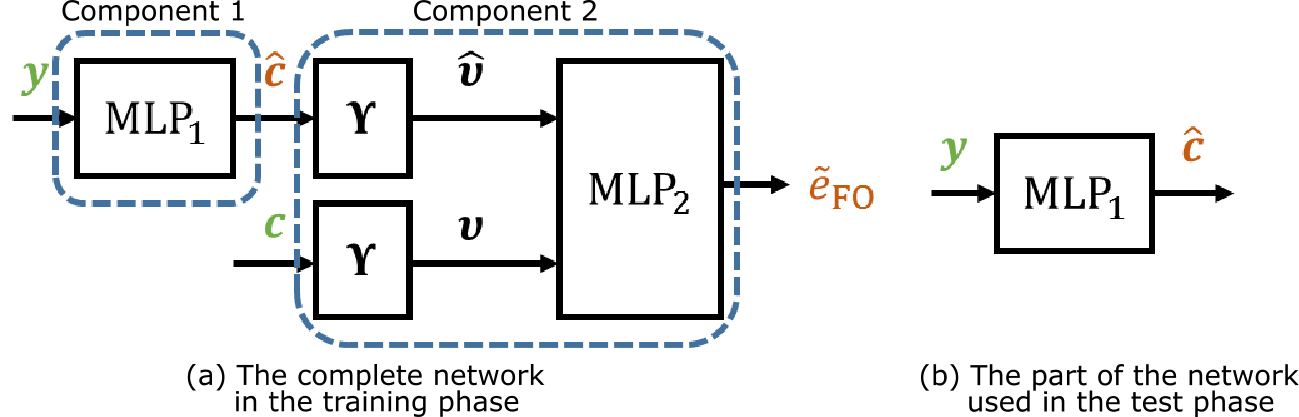}
\caption{The proposed deep network for EAP estimation: (a) the complete network in the training phase and (b) the part used in the test phase for EAP estimation. The input and output of the network are indicated by green and orange, respectively.}
\label{fig:dn}
\end{figure}

In this work, we propose a deep network structure described in Fig.~\ref{fig:dn}(a), which uses the error of FOs as an auxiliary variable to regularize the computation of the EAP in the training phase. The network comprises two major components: 1) an MLP ($\mathrm{MLP}_{1}$) that transforms the normalized diffusion signals $\bm{y}$ to the basis coefficients that represent the EAP; and 2) the mapping $\mathbf{\Upsilon}$ to compute ODF values in sampled directions for the estimated EAP $\hat{\bm{c}}$ and the training EAP $\bm{c}$, followed by another MLP ($\mathrm{MLP}_{2}$) that approximates the errors of FO estimation using the estimated ODF $\hat{\bm{v}}$ and the ODF $\bm{v}$ computed from the training EAP. The approximated FO error is denoted by $\tilde{\bm{e}}_{\mathrm{FO}}$.
Note that $\mathbf{\Upsilon}$ is known when the sampled directions are determined and is not trainable; and here we use 100 sampled directions. 
The second MLP is used to approximate the FO error because direct computation of the FO error from the ODF consists of complicated steps and the gradient of the error is difficult to compute during the training process. In this way, the gradient of the FO error can be easily computed for training the network. 
Like in~\cite{Golkov}, each MLP has three hidden layers. In each hidden layer, we use 500 units, which are more than the number used in~\cite{Golkov}, because we observed that EAP computation requires higher expressive power of the MLP.
In the test phase, only the first component of the network is needed for EAP computation, as shown in Fig.~\ref{fig:dn}(b).

\subsection{Training and Evaluation}

The two MLPs were trained separately in three steps as follows.
First, we trained the first component ($\mathrm{MLP}_{1}$) individually using only the mean squared errors of the EAP coefficients as the loss function. This step gives initial estimates of the EAPs and ODFs of the training samples, which are computed from the training diffusion signals. Second, using the estimated ODFs in the first step, the ODFs directly computed from the training EAPs, and the FO errors between these ODFs (computed using their peaks and the error measure in~\cite{FORNI}), we trained the second MLP to approximate the FO error computation, where the mean squared errors of FO errors were used as the loss function.
Finally, we fixed the second MLP and trained the entire network by minimizing the weighted sum of the approximated FO errors and mean squared errors of EAP coefficients
\begin{eqnarray}
\mathcal{L} = \sum_{i}\left(\alpha||\hat{\bm{c}}_{i}-\bm{c}_{i}||_{2}^{2} + \tilde{\bm{e}}_{\mathrm{FO},i}\right),
\label{eqn:obj}
\end{eqnarray} 
where $i$ is the index of training samples and $\alpha$ is a weighting constant (set to $\alpha=0.5$ empirically).
In each training step, we used the Adam algorithm~\cite{Kingma} as the optimizer, where the learning rate was 0.001  and the batch size was 128, and 10\% of the training samples were used as a validation set to prevent overfitting. The number of epochs was~10 for the first and final training steps and 40 for the second step. 
The network was implemented using Keras~({\url{http://keras.io/}}).

We used a strategy similar to~\cite{Golkov} to produce training data. 
For each dataset of dMRI scans that are acquired with a fixed set $\mathcal{G}$ of diffusion gradients, one deep network is trained.
To acquire training EAPs (and thus the diffusion features computed from them), training dMRI scans need to be acquired with a set $\tilde{\mathcal{G}}$ of diffusion gradients. $\tilde{\mathcal{G}}$ should densely sample the $q$-space, and satisfy $\mathcal{G}\subseteq\tilde{\mathcal{G}}$. 
Then, each voxel in the training scans is a training sample. The EAPs estimated by the conventional SHORE method using the complete diffusion gradients $\tilde{\mathcal{G}}$ are the training EAPs, which were used to train the network together with the normalized diffusion signals associated with the sparser diffusion gradients~$\mathcal{G}$. Here, we selected the implementation of SHORE in the Dipy software~\cite{DIPY} and used the default parameters ($(N,\zeta,\lambda_{N},\lambda_{L})=(6,700,10^{-8},10^{-8})$) provided at~\url{http://nipy.org/dipy/examples_built/reconst_shore.html}.

Similar to the training data generation, for evaluation on each test dMRI scan acquired with diffusion gradients~$\mathcal{G}$, diffusion gradients $\tilde{\mathcal{G}}$ densely sampling the $q$-space were applied. Gold standard EAPs were computed using $\tilde{\mathcal{G}}$ and the conventional estimation.
To evaluate the quality of EAP estimation, representative diffusion properties extracted from the estimated and gold standard EAPs were compared, including scalar features of \textit{mean square displacement}~(MSD)~\cite{Wu} and \textit{return-to-the-origin probability}~(RTOP)~\cite{Ozarslan2013}, and the directional feature of FOs. For the scalar quantities, we measured the mean absolute difference of MSD and the mean absolute difference of the cube root of RTOP (denoted by $\mathrm{RTOP}^{1/3}$)~\cite{Ozarslan2013}. MSD is related to the mean diffusivity~\cite{Wu} and RTOP represents the reciprocal of the statistical mean pore volume~\cite{Ozarslan2013}. For the directional information, we computed the difference of FOs using the error measure in~\cite{FORNI}. 

\section{Results}
\label{sec:exp}

We randomly selected brain dMRI scans of 10 subjects from the Human Connectome Project~(HCP) dataset~\cite{VanEssen}. The ten subjects were equally divided into two groups and a two-fold cross validation was used for evaluation.
The \textit{diffusion weighted images}~(DWIs) were acquired on a 3T MR scanner, where three shells ($b$-values of 1000, 2000, and 3000~$\mathrm{s}/\mathrm{mm}^{2}$) were used. Each $b$-value is associated with 90 gradient directions. The image resolution is 1.25~mm isotropic. 
For demonstration of the proposed method, we selected 60 fixed diffusion gradients as the diffusion gradients $\mathcal{G}$ for each training or test scan. These diffusion gradients consist of two shells ($b=1000,2000~\mathrm{s}/\mathrm{mm}^{2}$), each having 30 gradient directions, and resemble clinically achievable diffusion gradients. 
The training and gold standard EAPs were computed with all 270 diffusion gradients; and the normalized diffusion signals corresponding to $\mathcal{G}$ form the input $\bm{y}$ to the network. The training took about 8 hours on a 8-core Linux machine and used about 3,000,000 voxels.

\begin{figure}[!t]
  \centering
	\includegraphics[width=0.7\columnwidth]{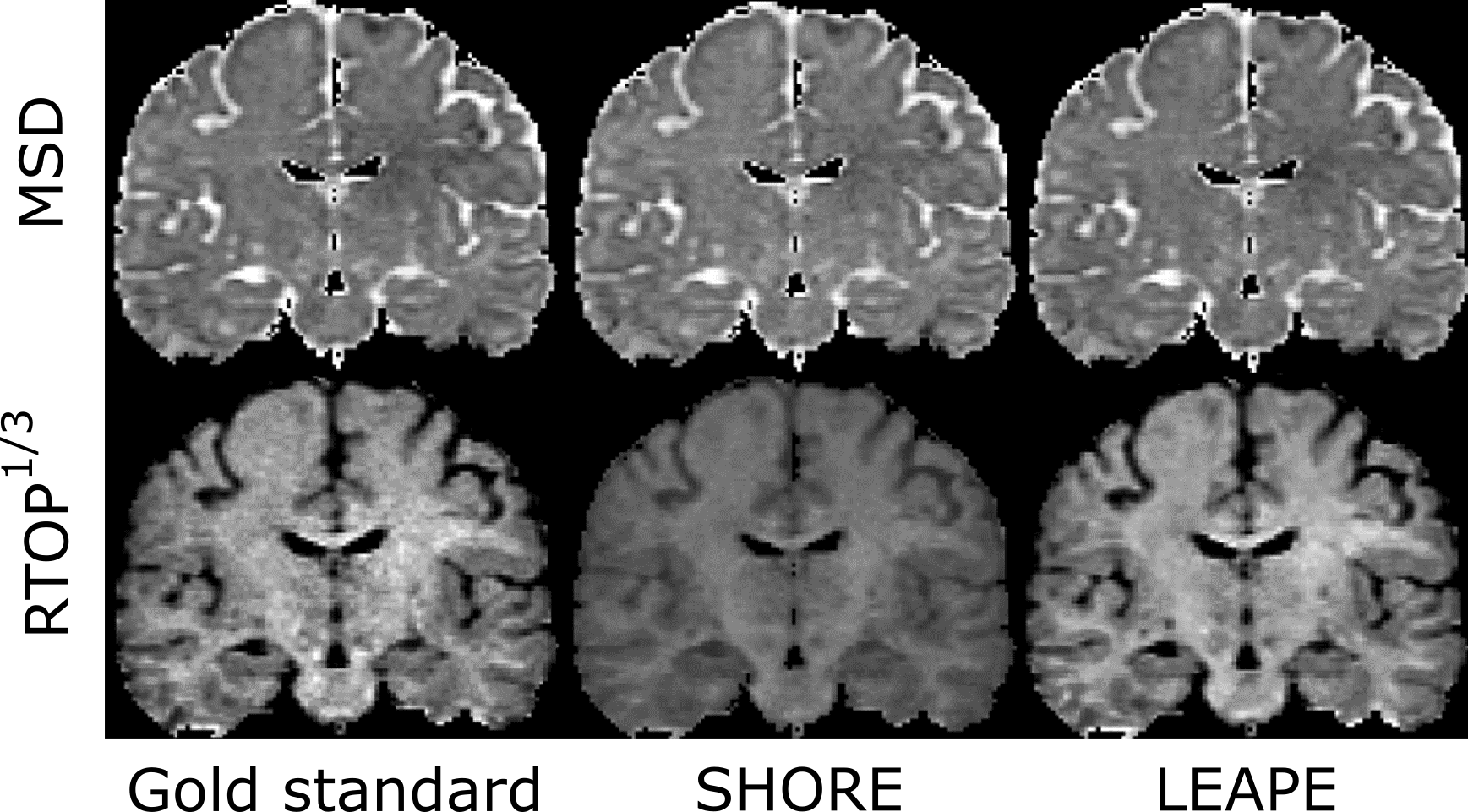}
\caption{Qualitative evaluation of MSD and RTOP in a representative coronal slice.}
\label{fig:scalar}
\end{figure}

\begin{figure}[!t]
  \centering
	\includegraphics[width=0.7\columnwidth]{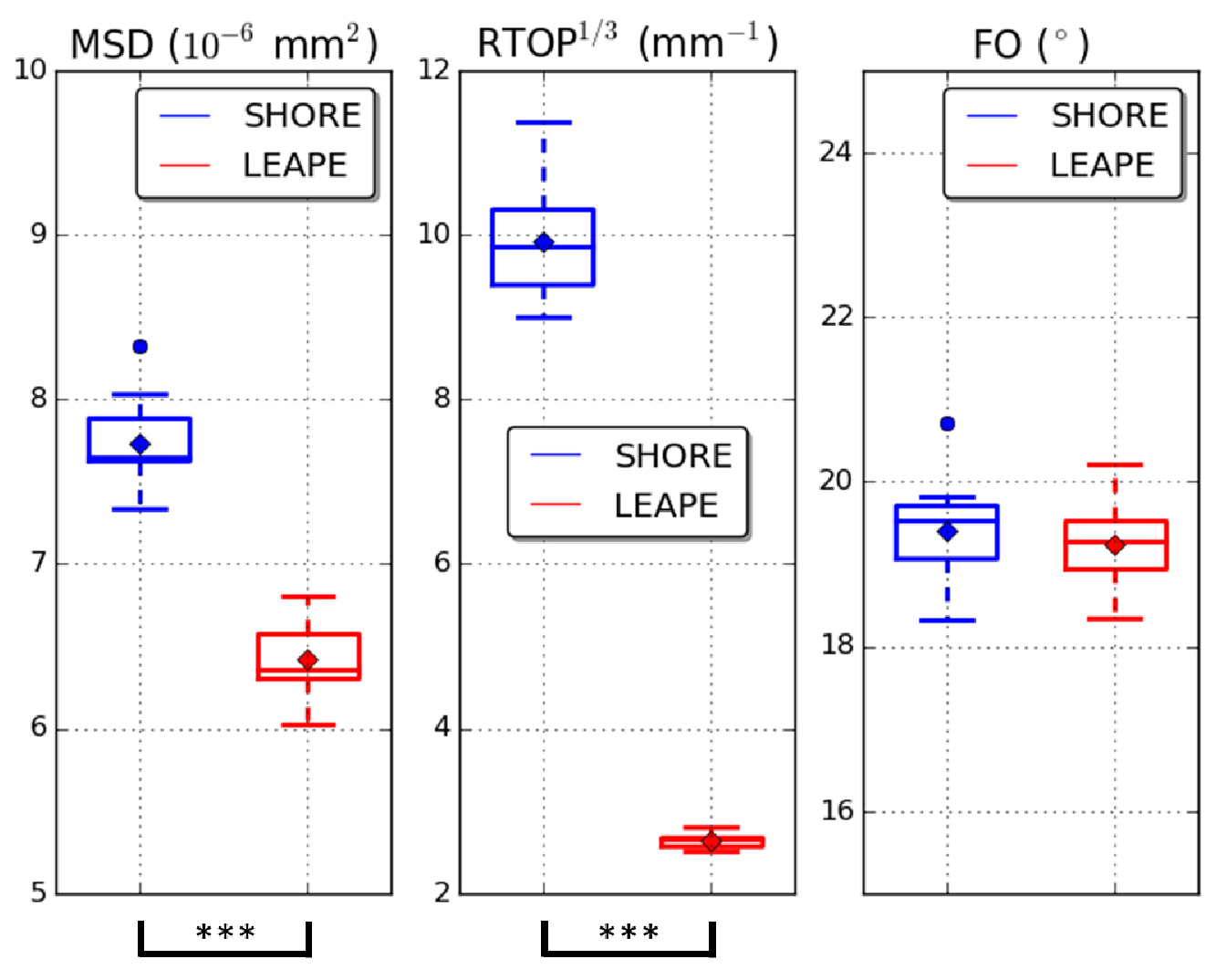}
\caption{Boxplots of the quantitative disagreement between the estimated results (MSD, RTOP, and FOs) and the gold standard for the ten subjects. The means are indicated by the diamonds. Asterisks (***) indicate that the difference between the inaccuracy of LEAPE and SHORE is highly significant ($p<0.001$) using a paired Student's $t$-test.}
\label{fig:diff}
\end{figure}

First, we evaluated the EAP estimation by examining the scalar features: MSD and RTOP. The LEAPE results on a representative subject are shown and compared with the gold standard and the results obtained by the conventional SHORE method~\cite{Merlet} in a coronal slice in Fig.~\ref{fig:scalar}. Note that here for each MSD or $\mathrm{RTOP}^{1/3}$ map, the color map is the same for the three columns. The LEAPE results resemble the gold standard. The RTOP result of SHORE is remarkably biased with the smaller set $\mathcal{G}$ of diffusion gradients. Then, we computed the average disagreement in the brain between the estimates and gold standard for the ten subjects, which is shown in the boxplots in Fig.~\ref{fig:diff}. For both MSD and $\mathrm{RTOP}^{1/3}$, the LEAPE errors are smaller than those of SHORE, and the differences are highly significant ($p<0.001$) using a paired Student's $t$-test.

\begin{figure}[!t]
  \centering
	\includegraphics[width=0.7\columnwidth]{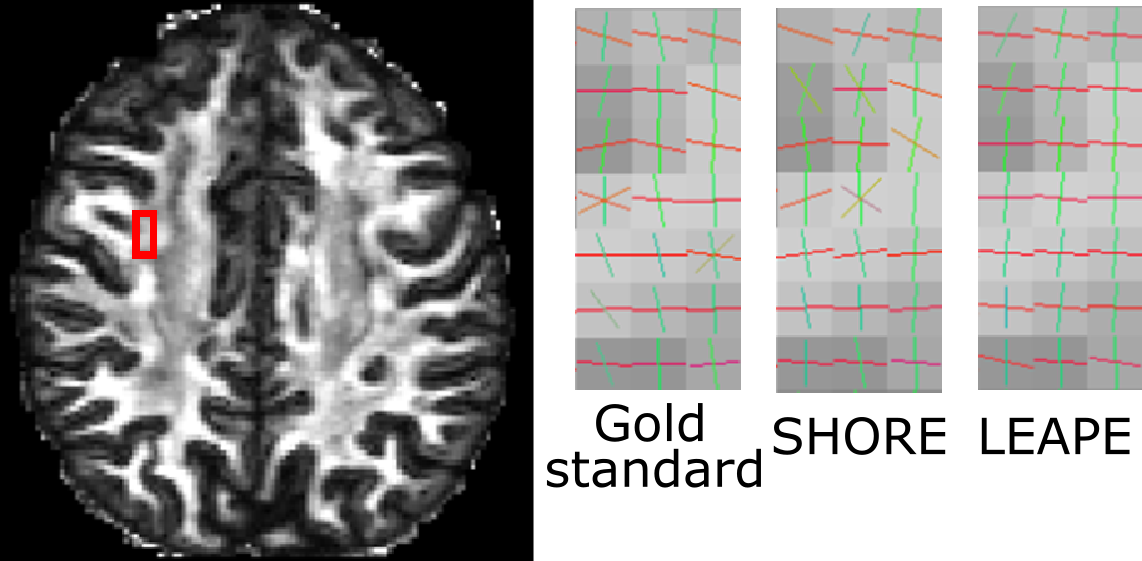}
\caption{FOs overlaid on the FA map in a region where the CC and SLF cross.}
\label{fig:directional}
\end{figure}

Next, we evaluated the FOs extracted from the EAP. A qualitative comparison is made in Fig.~\ref{fig:directional}, where we focus on a region where the \textit{corpus callosum}~(CC) and \textit{superior longitudinal fasciculus}~(SLF) cross. We can see that the LEAPE result resembles the gold standard, and LEAPE better resolves crossing FOs and produces smoother FOs than SHORE. The LEAPE FOs are even smoother than the gold standard in some cases. We also computed the average FO disagreement with the gold standard in the white matter for SHORE and LEAPE (see Fig.~\ref{fig:diff}). 
Note that if the second MLP in LEAPE is not used for training, the FO errors range from $20^{\circ}$ to $24^{\circ}$ (not shown in Fig.~\ref{fig:diff}), and for every subject the errors are higher ($>10\%$) than those of SHORE or LEAPE. This indicates the benefit of adding the second MLP.
Compared with SHORE, the mean and median of the disagreement of LEAPE are smaller. The difference of disagreement between SHORE and LEAPE is small. This is possibly because for the less complicated FO configurations (for example, noncrossing FOs), which occupy a large proportion of the white matter volume, both SHORE and LEAPE are able to produce good results with 60 diffusion gradients, and the difference like the one shown in Fig.~\ref{fig:directional} is present at regions with more complex FO configurations. 

\section{Conclusion}
\label{sec:conclusion}

We have proposed a deep network to estimate the EAP, which comprises two cascaded components. The first component uses an MLP to perform EAP estimation, and the second one approximates FO errors with the aid of a second MLP to regularize the training. The proposed method was applied to real brain dMRI and results demonstrate improved estimation of diffusion features compared with the conventional EAP estimation approach.

\section*{Acknowledgement}

This work is supported by NSFC 61601461.
Data were provided by the Human Connectome Project, WU-Minn Consortium (Principal Investigators: David Van Essen and Kamil Ugurbil; 1U54MH091657).

\bibliographystyle{splncs03}
\bibliography{refs}

\begin{thebibliography}{10}
\providecommand{\url}[1]{\texttt{#1}}
\providecommand{\urlprefix}{URL }

\bibitem{Assemlal}
Assemlal, H.E., Tschumperl{\'e}, D., Brun, L.: {Efficient and robust
  computation of PDF features from diffusion MR signal}. Medical Image Analysis
   13(5),  715--729 (2009)

\bibitem{Descoteaux}
Descoteaux, M., Deriche, R., Le~Bihan, D., Mangin, J.F., Poupon, C.: Multiple
  q-shell diffusion propagator imaging. Medical Image Analysis  15(4),
  603--621 (2011)

\bibitem{DIPY}
Garyfallidis, E., Brett, M., Amirbekian, B., Rokem, A., Van Der~Walt, S.,
  Descoteaux, M., Nimmo-Smith, I., Contributors, D.: Dipy, a library for the
  analysis of diffusion {MRI} data. Frontiers in Neuroinformatics  8(8),  1--17
  (2014)

\bibitem{Golkov}
Golkov, V., Dosovitskiy, A., Sperl, J.I., Menzel, M.I., Czisch, M., S{\"a}mann,
  P., Brox, T., Cremers, D.: q-space deep learning: Twelve-fold shorter and
  model-free diffusion {MRI} scans. IEEE Transactions on Medical Imaging
  35(5),  1344--1351 (2016)

\bibitem{johansen}
Johansen-Berg, H., Behrens, T.E.J.: Diffusion {MRI}: from quantitative
  measurement to in vivo neuroanatomy. Waltham: Academic Press (2013)

\bibitem{Kingma}
Kingma, D., Ba, J.: Adam: A method for stochastic optimization. arXiv preprint
  arXiv:1412.6980  (2014)

\bibitem{LeCun}
LeCun, Y., Bengio, Y., Hinton, G.: Deep learning. Nature  521(7553),  436--444
  (2015)

\bibitem{Merlet}
Merlet, S.L., Deriche, R.: {Continuous diffusion signal, EAP and ODF estimation
  via Compressive Sensing in diffusion MRI}. Medical Image Analysis  17(5),
  556--572 (2013)

\bibitem{Ozarslan2013}
{\"O}zarslan, E., Koay, C.G., Shepherd, T.M., Komlosh, M.E.,
  {\.I}rfano{\u{g}}lu, M.O., Pierpaoli, C., Basser, P.J.: {Mean apparent
  propagator (MAP) MRI: a novel diffusion imaging method for mapping tissue
  microstructure}. NeuroImage  78,  16--32 (2013)

\bibitem{VanEssen}
Van~Essen, D.C., Smith, S.M., Barch, D.M., Behrens, T.E.J., Yacoub, E.,
  Ugurbil, K.: The {WU-Minn} human connectome project: An overview. NeuroImage
  80(0),  62--79 (2013)

\bibitem{Wu}
Wu, Y.C., Field, A.S., Alexander, A.L.: Computation of diffusion function
  measures in $q$-space using magnetic resonance hybrid diffusion imaging. IEEE
  Transactions on Medical Imaging  27(6),  858--865 (2008)

\bibitem{FORNI}
Ye, C., Zhuo, J., Gullapalli, R.P., Prince, J.L.: Estimation of fiber
  orientations using neighborhood information. Medical Image Analysis  32,
  243--256 (2016)

\end{thebibliography}
\end{document}